\documentclass[letterpaper, 10 pt, conference]{ieeeconf}  

\IEEEoverridecommandlockouts                              

\usepackage{times}  
\usepackage{helvet}  
\usepackage{courier}  
\usepackage[hyphens]{url}  
\usepackage{graphicx} 

\usepackage{caption} 

\usepackage{algorithm}
\usepackage{algorithmic}

\usepackage{newfloat}
\usepackage{listings}
\floatstyle{ruled}
\newfloat{listing}{tb}{lst}{}
\floatname{listing}{Listing}
\usepackage{booktabs} 
\usepackage{amsmath}
\usepackage{amssymb}
\usepackage{mathtools}

\usepackage[capitalize,noabbrev]{cleveref}

\usepackage{soul}
\usepackage{multirow}
\usepackage{caption}
\usepackage[table]{xcolor}

\usepackage{amsfonts}       
\usepackage{nicefrac}       
\usepackage{microtype}      
\usepackage{textcomp}
\usepackage{times}
\usepackage{epsfig}
\usepackage{lipsum}
\usepackage{tabularx}

\newcommand{\first}[1]{\cellcolor{cyan!20}\textbf{#1}}
\newcommand{\second}[1]{\cellcolor{yellow!20}#1}

\usepackage{listings}

\usepackage{xspace}

\newif\ifhighlightchanges
\highlightchangesfalse   
\ifhighlightchanges
  \newcommand{\rev}[1]{\textcolor{blue}{#1}}
\else
  \newcommand{\rev}[1]{#1}
\fi

\title{
FastMap: Real-Time Semantic Map Completion via Bitwise Masked Modeling
}

\author{Yijie Deng$^{1,2,3}$, Shuaihang Yuan$^\dagger$$^{1,2,3,4}$, Congcong Wen$^{1,2,4}$, Hao Huang$^{1,2,4}$ and Anthony Tzes$^{1,2}$
\thanks{$^\dagger$ Corresponding Author.}
\thanks{$^{1}$NYUAD Center for Artificial Intelligence and Robotics, Abu Dhabi, UAE.}%
\thanks{$^{2}$New York University Abu Dhabi, Electrical Engineering, Abu Dhabi 129188, UAE.}%
\thanks{$^{3}$New York University, Electrical \& Computer Engineering Dept., Brooklyn, NY 11201, USA.}
\thanks{$^{4}$Embodied AI and Robotics (AIR) Lab, NYU Abu Dhabi, UAE.}
}

\begin{document}
\bstctlcite{FastMap:BSTcontrol}

\maketitle
\thispagestyle{empty}
\pagestyle{empty}


\begin{abstract}
Semantic map completion, which predicts the layout of unobserved regions
from partial observations, is a critical capability for indoor
robot navigation. Existing approaches either rely on high-dimensional
discrete codebooks that inflate memory, or on iterative diffusion
sampling that is too slow for real-time use. We present
\textbf{FastMap}, a lightweight two-stage framework for completing
top-down categorical semantic maps. First, a lookup-free BitVAE
exploits the inherently binary (one-hot) structure of semantic maps to
compress each map patch into compact bitwise tokens, yielding a 0.41\,GB
model that is 3.7$\times$ smaller than the prior masked-modeling baseline.
Second, a Masked AutoEncoder (MAE)-style transformer reconstructs
missing tokens in a single forward pass at
0.011\,s/map. To support object goal navigation, we additionally
introduce an object-aware masking strategy that masks the target
category during training and conditions generation on a learnable
category embedding, without adding inference cost. On the Gibson
benchmark, FastMap achieves 34.10\% mIoU and 45.84\% semantic
Success Rate (sSR), more than doubling the previous best
(21.88\%), and reaches 83.8\% Success Rate on downstream
ObjectNav, \rev{the highest among the compared navigation methods}.
\end{abstract}


\section{Introduction}

Robots in cluttered indoor environments must not only recognize what is directly visible, but also anticipate the semantics of unobserved space. Standard semantic mapping accumulates per-frame predictions into a partial top-down map, leaving large regions unknown under limited viewpoints. For tasks such as object goal navigation, reasoning about these unobserved regions can reduce unnecessary exploration and improve success rates.

Learning priors over indoor semantic layouts is challenging: top-down semantic maps are sparse, class-imbalanced, and contain objects at vastly different scales. Existing approaches \rev{fall into two broad categories}. Completion-based methods predict missing regions from the current partial map using CNN- or GNN-based architectures~\cite{liang2021sscnav,sun2024rsmpnet}, while generative approaches use masked modeling or diffusion to hallucinate unseen areas, sometimes paired with LLMs for planning~\cite{zhang2024imagine,ji2024diffusion}. These pipelines are often memory-heavy and slow, which hinders real-time deployment. \rev{Navigation-centric systems further embed imagination modules in a full stack: ForesightNav~\cite{shah2025foresightnav} predicts occupancy and CLIP features on a dense GeoSem map, GOAL~\cite{li2025distilling} trains an LLM-distilled flow model, and NaviFormer~\cite{xie2025naviformer} fuses multi-source top-down maps with a spatio-temporal transformer.} These works raise a practical question: how can we build a lightweight, plug-and-play completion module that runs in real time on resource-constrained robots?

\begin{figure*}[t]
    \centering
    \includegraphics[width=0.74\textwidth]{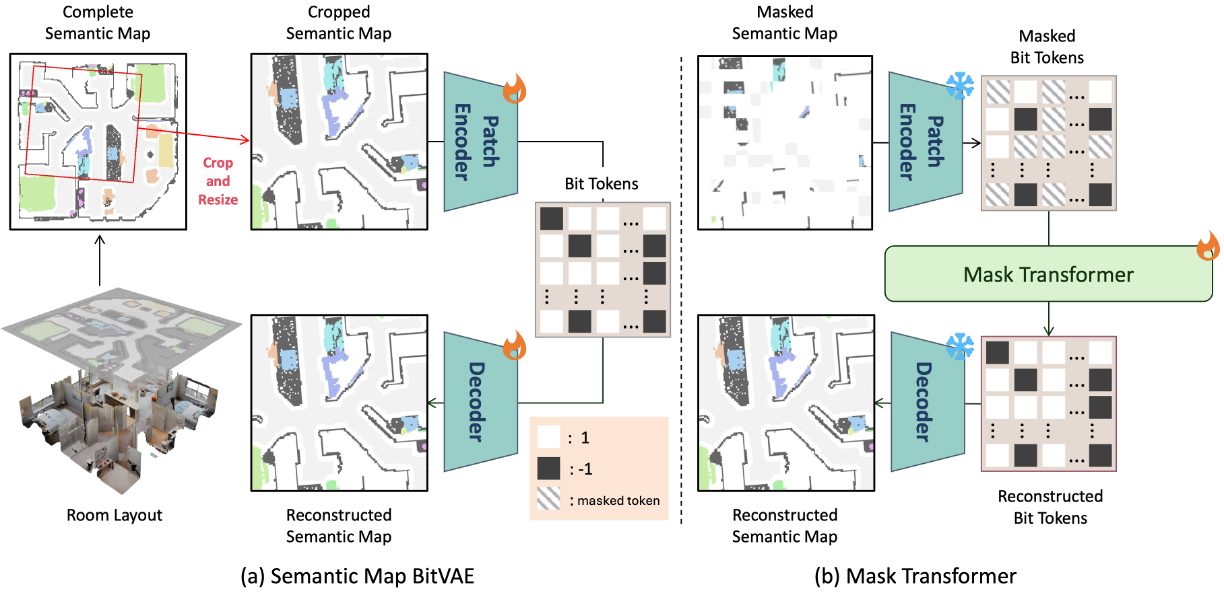}
    \caption{Overview of FastMap. The pipeline consists of two stages: (1) a lookup-free BitVAE that learns compact bitwise tokens for semantic maps, and (2) a masked transformer that predicts missing tokens from partial observations and decodes a complete semantic map.}
    \label{fig:method_overview}
\end{figure*}

To balance spatial reasoning and efficiency, we propose FastMap, a lightweight framework for completing top-down semantic maps from partial observations. \rev{As illustrated in Figure~\ref{fig:method_overview},} FastMap uses a two-stage pipeline. First, motivated by the one-hot nature of semantic maps, we encode each map patch into a compact bitwise token using a lookup-free BitVAE. Second, we train a BERT-style masked transformer to predict missing tokens and reconstruct a complete semantic map.

To improve object-centric reasoning for navigation, we introduce an object-aware masking strategy: instead of masking patches uniformly at random, we additionally mask all patches containing the queried goal category and condition the transformer on a learnable category embedding. \rev{This teaches a prior over likely target locations without adding any inference-time computation.}

The contributions of our method are summarized as follows:
\begin{enumerate}
\item \textbf{FastMap: real-time semantic map completion.} An efficient, plug-and-play framework that encodes semantic maps into compact bitwise tokens via a lookup-free BitVAE and restores missing regions with a BERT-style masked transformer.
\item \textbf{Target-conditioned object-aware masking.} Unlike conventional random patch masking, our approach masks an entire goal category and adds a learnable category embedding, enhancing object-centric reasoning without extra inference cost.
\item \textbf{Strong performance with low latency.} FastMap attains state-of-the-art map generation quality and improves object goal navigation on Gibson while running at 0.011\,s per map with a 0.41\,GB model.
\end{enumerate}

\section{Related Work}
Semantic map completion for robot navigation sits at the intersection of two
threads: (i) learning-based spatial reasoning that predicts missing map regions
from partial observations, and (ii) generative masked modeling that reconstructs
corrupted discrete token sequences. We review both and explain how FastMap builds
upon them.

\subsection{Semantic Map Reasoning}
\label{sec:rw_smg}
Semantic maps embed object-level semantics in a top-down layout and are widely used for embodied navigation. Classical systems aggregate per-frame segmentation into a global map (e.g., SemanticFusion~\cite{mccormac2017semanticfusion}), leaving large regions unobserved when exploration is limited. \rev{To reason about these regions, completion-based methods inpaint the partial map: SSCNav~\cite{liang2021sscnav} uses a confidence-aware CNN and RSMPNet~\cite{sun2024rsmpnet} relationship-guided graph reasoning, while others predict \emph{where} unseen targets lie rather than reconstructing the map, e.g., PEANUT~\cite{zhai2023peanut} regresses a target-location probability map. Such methods are typically tied to a specific policy and predict occupancy or a scalar target prior rather than a full categorical layout.}

More recently, generative semantic mapping aims to hallucinate plausible unobserved semantics by learning priors from complete maps. SGM~\cite{zhang2024imagine} applies masked modeling to generate semantic maps for object goal navigation, but relies on high-dimensional discrete tokens, leading to a substantial memory footprint. \rev{Diffusion-based reasoning~\cite{ji2024diffusion} instead denoises a full semantic map with LLM-biased guidance, at the cost of iterative sampling.} In contrast, our work focuses on making masked map generation efficient by using bitwise tokens and lightweight transformers, while maintaining strong object-centric reasoning through target-aware training.

Recent navigation systems also embed imagination modules with different representations: ForesightNav~\cite{shah2025foresightnav} stores high-dimensional CLIP features, GOAL~\cite{li2025distilling} relies on iterative flow sampling, and NaviFormer~\cite{xie2025naviformer} fuses online top-down maps without explicitly completing unseen semantics. FastMap is complementary: it completes categorical semantic maps in real time with a compact bitwise representation that plugs into diverse pipelines.

\subsection{Generative Masked Modeling}
\label{sec:rw_gmm}
Most masked generative pipelines rely on a discrete tokenizer learned with a codebook (e.g., VQ-VAE~\cite{van2017vqvae}), while recent work shows that embedding-free bit tokens can remove codebook lookups and improve efficiency~\cite{weber2024maskbit}. 

BERT~\cite{devlin2019bert} pioneered masked language modeling, and MaskGIT~\cite{chang2022maskgit} and MAGE~\cite{li2023mage} extend it to image generation with iterative decoding; later works adapt masked modeling to image, video, and motion generation~\cite{guo2024momask,pinyoanuntapong2024mmm}. \rev{Whereas prior map-completion networks operate in pixel space and recent generative approaches use heavy learned tokens, FastMap extends efficient bitwise masked modeling to indoor semantic map completion for real-time navigation.}

\section{Method}


\subsection{Problem Formulation}
\label{sec:problem_formulation}

We aim to generate complete semantic maps from partial observations. Let $M \in \mathbb{R}^{H \times W \times C}$ be a complete one-hot semantic map ($H,W$ spatial dimensions, $C$ categories) and $M_{\text{partial}}$ a partial observation with unobserved regions masked to zero. \rev{$M_{\text{partial}}$ may come from various sources (e.g., RGB-D with semantic segmentation); our method is agnostic to the source.} We learn a generative function $f: M_{\text{partial}} \rightarrow \hat{M}$ that predicts the complete map while capturing the distribution of layouts in unobserved areas, \rev{enabling plausible completions that respect the structural and semantic regularities of indoor scenes.}


\subsection{Learning Map Representations with BitVAE}
\label{sec:vqvae}

\begin{figure}[t]
    \centering
    \begin{minipage}[b]{0.49\linewidth}
        \centering
        \includegraphics[width=\linewidth]{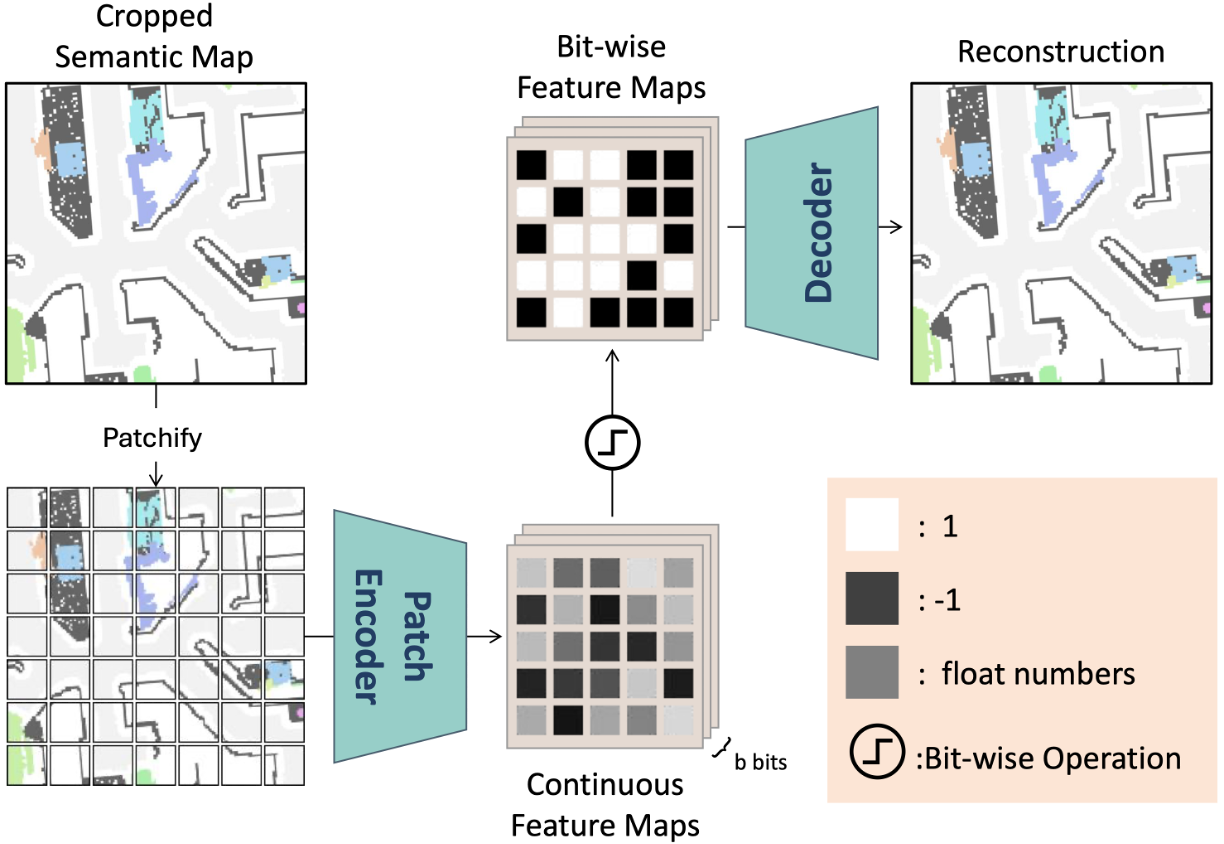}
        \caption{Overview of the BitVAE map reconstruction process.}
        \label{fig:bitvae}
    \end{minipage}\hfill
    \begin{minipage}[b]{0.49\linewidth}
        \centering
        \includegraphics[width=\linewidth]{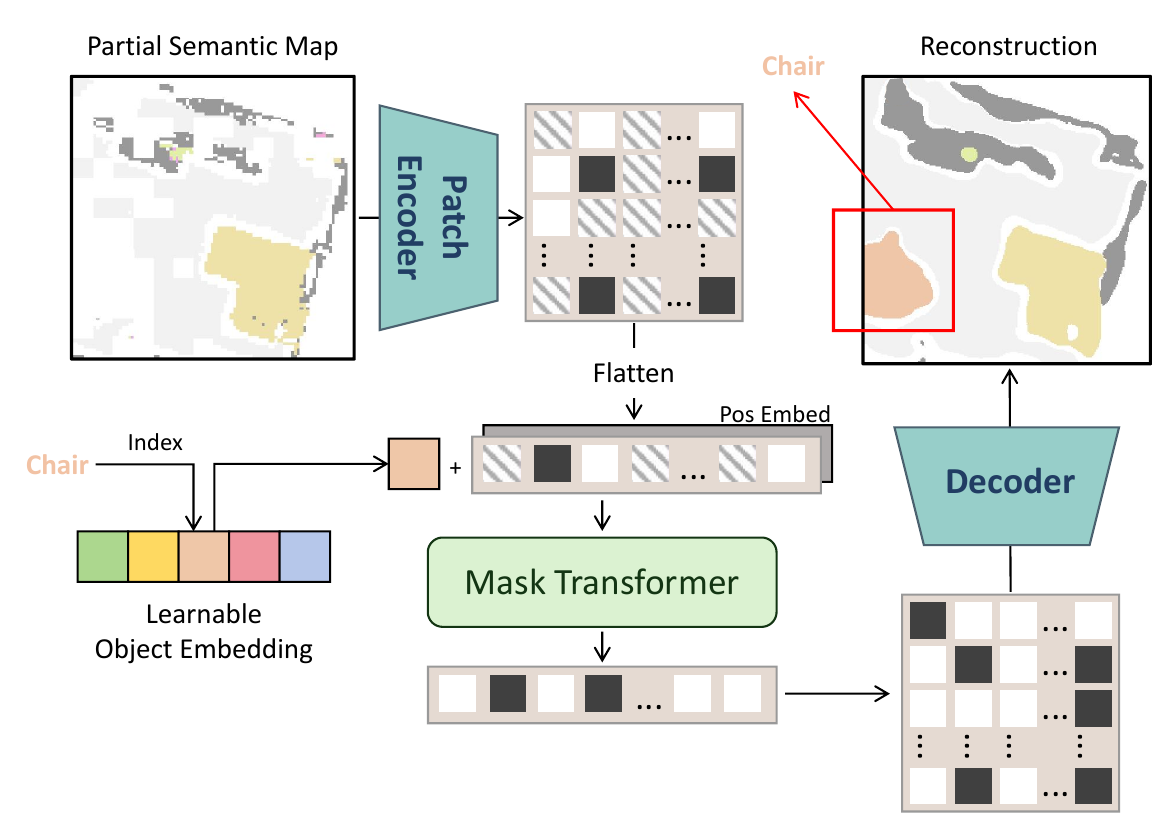}
        \caption{Overview of the mask transformer generation process.}
        \label{fig:mask_transformer}
    \end{minipage}
\end{figure}

To effectively represent indoor semantic maps, we employ the BitVAE architecture~\cite{weber2024maskbit} that encodes semantic maps into a binary discrete latent space. Traditional BERT-like masked modeling approaches~\cite{guo2024momask,pinyoanuntapong2024mmm,chen2024don} typically use a discrete codebook, which requires a lookup table of size $N \times C_d$ ($N$ codes of dimension $C_d$). While codebooks excel at information-dense signals such as color images and motion sequences, they are suboptimal for indoor semantic maps: our key insight is that such maps are inherently sparse in their one-hot representation, containing only binary values $0$ (absence) and $1$ (presence), which makes BitVAE particularly well-suited. The encoding process, illustrated in Figure~\ref{fig:bitvae}, proceeds as follows:

A patch encoder $E(\cdot)$~\cite{he2022masked} first transforms the semantic map $M \in \mathbb{R}^{H \times W \times C}$ into a feature map $m = E(M) \in \mathbb{R}^{h \times w \times b}$, where $b$ is the number of encoding bits and $h,w$ the feature-map size. The feature map is then binarized by $B(\cdot)$:
   \begin{equation}
   B(m_{i,j,k}) = 
   \begin{cases}
   1, & \text{if } m_{i,j,k} > 0 \\
   -1, & \text{otherwise}
   \end{cases}
   \end{equation}
   where $m_{i,j,k}$ is the $k$-th bit of the bit-wise feature map $m$ at position $(i,j)$.

The resulting binary feature representation $B(m) \in \{-1, 1\}^{h \times w \times b}$ is then processed by a CNN-based decoder $D(\cdot)$ to reconstruct the semantic map:
   \begin{equation}
   \hat{M} = D(B(m))
   \end{equation}

\rev{This BitVAE approach offers two advantages: (1) improved reconstruction fidelity from the natural alignment between the binary latent and the one-hot map structure, and (2) enhanced compatibility with the mask transformer's token restoration. We validate its effectiveness in Section~\ref{sec:exp_ablation}.}

\noindent\textbf{Scaling with categories and bits.} \rev{The BitVAE latent has shape $h\times w\times b$, so memory grows linearly with the bits $b$ and grid size $h\times w$; the category count $C$ only affects the encoder/decoder channels (no exponential dependence), and the transformer predicts from a vocabulary of $(2^b{+}1)$. In contrast, modules storing dense per-cell embeddings (e.g., CLIP with $D{=}512$~\cite{shah2025foresightnav}) scale linearly with $D$ at full resolution.}

\subsection{Map Generation with Mask Transformer}
\label{sec:mask_transformer}

After training, we use the BitVAE as a discrete tokenizer. Given a partial semantic map $M_{\text{partial}}$ aggregated from the agent's observations, the encoder $E(\cdot)$ produces bit indices $I_{\text{partial}}\in[0, 2^{b}-1]$ over the observed area, and unobserved regions are assigned a special mask index $2^b$. The Mask Transformer follows BERT~\cite{devlin2019bert}, using bidirectional self-attention to predict the complete indices $I_{\text{complete}}$ from the observed context, \rev{capturing both local geometric constraints and global semantic patterns}.

During training, we optimize the Mask Transformer by minimizing the cross-entropy loss between predicted and ground-truth indices at masked positions:
\begin{equation}
    \mathcal{L}_{\text{MT}} = -\sum_{i,j} \log p(I_{\text{complete}}^{(i,j)} | I_{\text{partial}})
\end{equation}
where $p(\cdot | I_{\text{partial}})$ is the predicted distribution over indices at position $(i,j)$ given the partial observation.

\subsection{Target-conditioned Map Completion for ObjectNav}
\label{sec:extension}

In object goal navigation, the agent is given a target object category $c \in \{1,...,n\}$ to locate ($n$ categories in total). We leverage this target to enhance the transformer's object-location prediction: as illustrated in Figure~\ref{fig:mask_transformer}, we maintain a learnable target embedding $E \in \mathbb{R}^{n \times d}$ with embedding dimension $d$.

\rev{During navigation, the goal category $c$ is given by the ObjectNav task specification, so the category embedding serves purely as a conditioning signal. If multiple categories are queried, FastMap can be invoked once per category and the predictions merged by per-pixel argmax. FastMap can also run unconditionally by omitting the category token, corresponding to our ``w.o. ObMask'' variant.}

During training we apply two complementary masking strategies to each map $M \in \mathbb{R}^{H \times W \times C}$: (i) random masking of a patch subset $\mathcal{P}_r$ for general scene understanding, and (ii) target-specific masking that selects a category $c$ and masks all patches $\mathcal{P}_c$ where it appears ($M_{h,w,c}=1$). The masked map is encoded by BitVAE into tokens $B_{\text{masked}}$, concatenated with the target embedding $E_c$ to form $X=[B_{\text{masked}};E_c]$, and the transformer predicts the complete tokens $\text{MaskTransformer}(X)$ conditioned on both. \rev{This target-aware scheme develops a prior over likely locations of specific object categories, improving predictions during navigation.}





\subsection{Implementation and Training}
\label{sec:training}

We train FastMap in two procedures, detailed below. In summary, the semantic map is $224\times224$ (a room-layout channel plus 15 object channels); the BitVAE encoder is a patch CNN with stride 16 producing a $14\times14$ token grid with $b{=}9$ bits per token (vocabulary $2^b{+}1$ including the mask token); and the mask transformer is a ViT-B (12 layers, 12 heads, dim 768) with learned 2D positional embeddings and a 15-category goal embedding of dim 768.

\paragraph{BitVAE Training.} \rev{We resize each input semantic map to $224\times224$. The encoder is a patch-wise CNN with stride 16 that independently processes each patch into latent features, which are binarized into 9 bits per patch; the decoder is a residual CNN.} We train the BitVAE by minimizing a combination of two losses:

a \textbf{binary cross-entropy} loss between the one-hot map $M$ and the reconstruction $\hat{M}$, and a \textbf{map IoU} loss that directly optimizes segmentation quality:
\begin{equation}
\begin{split}
\mathcal{L}_{\text{BCE}} = -\frac{1}{HWC}\sum_{i,j,c} \big[ & M_{i,j,c} \log \hat{M}_{i,j,c} \\
& + (1-M_{i,j,c}) \log(1-\hat{M}_{i,j,c}) \big],
\end{split}
\end{equation}
\begin{equation}
\mathcal{L}_{\text{IoU}} = 1 - \frac{1}{C}\sum_{c=1}^C \frac{|M_c \cap \hat{M}_c|}{|M_c \cup \hat{M}_c|}.
\end{equation}
The total loss is $\mathcal{L}_{\text{BCE}}+\mathcal{L}_{\text{IoU}}$ (both weights set to 1.0).

\begin{figure*}[t]
  \centering
  \includegraphics[width=0.8\textwidth]{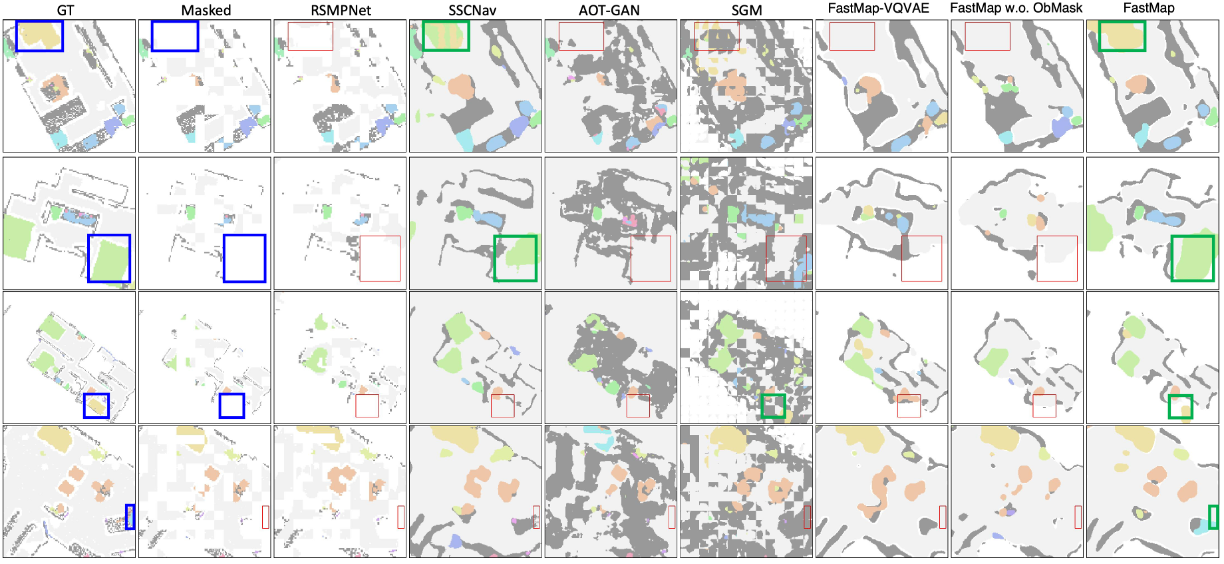}
  \caption{Qualitative comparison of map generation quality. The leftmost two columns show the ground truth semantic map and its masked version, with \textbf{blue} boxes highlighting the masked target objects using our object-aware masking strategy. \rev{This masked input is a controlled incomplete-exploration protocol for comparing how well methods recover \emph{unobserved} target objects (measured by sSR), rather than a realistic observation; the real navigation setting, where the agent progressively explores and pre-locates objects, is shown in Figure~\ref{fig:navigation}.} The subsequent columns display the generated results from different methods. Bold \textbf{green} boxes indicate successful target object localization, while \textbf{red} boxes denote failed attempts.}
  \label{fig:map_comparison}
\end{figure*}
\paragraph{Mask Transformer Training.} The mask transformer is a multi-layer multi-head self-attention network. Its input is the partial bit indices $I_{\text{partial}} \in \{0,\ldots,2^b-1\}^N$ from the BitVAE ($N$ patches, $b$ bits each), and it outputs logits $L \in \mathbb{R}^{M \times 2^b}$ over bit indices for the $M$ missing patches. We minimize the cross-entropy loss against the ground-truth indices $I_{\text{gt}}$:

\begin{equation}
\mathcal{L}_{\text{MT}} = -\frac{1}{M}\sum_{i=1}^M \log\left(\frac{\exp(L^{(i,I_{\text{gt}}^{(i)})})}{\sum_{j=0}^{2^b-1} \exp(L^{(i,j)})}\right)
\end{equation}

\rev{We employ a two-phase masking schedule: in the first quarter of training we mask a small portion $(15\text{-}20\%)$ of indices so the transformer learns global context, then gradually raise the mask ratio from $15\%$ to $75\%$ with cosine scheduling, following BERT-style practice~\cite{guo2024momask,pinyoanuntapong2024mmm}. Index $2^b$ is an input-only mask token, and the transformer predicts only valid token indices.}

\section{Experiment}

\subsection{Experimental Setup}
\label{sec:exp_setup}

\textbf{Dataset.} We evaluate FastMap on the Gibson indoor scenes~\cite{xia2018gibson}, following the splits of prior work~\cite{ramakrishnan2022poni,zhang20233d,zhang2024imagine}, with the standard 15 goal object categories and a room-layout channel. For map generation we follow SGM~\cite{zhang2024imagine} to obtain test data by random cropping/rotation of the ground-truth maps, and we use the standard Gibson ObjectNav splits and configuration of~\cite{zhang2024imagine, xie2025naviformer,li2025distilling}.

\textbf{Map generation metrics.} We report mean IoU (mIoU) over all channels and category-level Precision/Recall/F1 (a category counts as present if it occupies any pixel), \rev{which measure how faithfully the full map is reconstructed}. We also report \textbf{sSR} (semantic Success Rate), which asks whether the model can find a hidden object: we mask all pixels of a goal category $c$ (plus random patches) and count a success if it correctly predicts \emph{any} pixel of $c$ there (same protocol for all methods). \rev{sSR is thus a lenient, \emph{binary} check, so it reflects rough localization, not shape or completeness, and complements mIoU/F1. As it favors target-conditioned methods, we also report our unconditional ``FastMap w.o. ObMask'' variant for a fair comparison. }

\textbf{Navigation metrics.} For object goal navigation we follow the protocol of~\cite{zhang2024imagine,li2025distilling,xie2025naviformer} and report Habitat ObjectNav metrics: Success Rate (SR), Success weighted by Path Length (SPL), and Distance To Success (DTS). \rev{To isolate the map-completion module, we plug FastMap into the SGM~\cite{zhang2024imagine} stack by replacing its MAE-based generator, keeping the frontier planner, sensing (RGB-D with semantic segmentation), and data splits unchanged. Numbers for the other methods are taken from their papers under the standard Gibson setup; since these systems use different policies and training budgets, the comparison shows competitive downstream utility rather than a strictly controlled head-to-head result.}

\textbf{Baseline methods.} Following SGM~\cite{zhang2024imagine}, we compare against open-source RSMPNet~\cite{sun2024rsmpnet} (GNN), SSCNav~\cite{liang2021sscnav} (CNN), AOT-GAN~\cite{zeng2022aggregated} (inpainting), and SGM~\cite{zhang2024imagine} (masked modeling), plus two FastMap variants: FastMap-VQVAE (BitVAE replaced by a 512-code VQ-VAE) and FastMap w.o. ObMask (random masking, no category conditioning). FastMap receives the goal embedding for target-conditioned evaluation while baselines are unconditional, with w.o. ObMask as the fair unconditional comparison. We further evaluate FastMap as a plug-and-play module for ObjectNav~\cite{zhang2024imagine,xie2025naviformer,li2025distilling}.

\renewcommand{\arraystretch}{1.1}

\renewcommand{\arraystretch}{1.1}
\begin{table*}[t]
  \centering
  \caption{Comparison of map generation quality and model efficiency on the Gibson indoor scene~\cite{xia2018gibson} dataset.}
  \label{tab:combined}
  \resizebox{0.72\textwidth}{!}{%
  \begin{tabular}{lcccccccc}
    \toprule
    \multirow{2}{*}{Method} & \multicolumn{5}{c}{\textbf{Map Generation Quality}} & & \multicolumn{2}{c}{\textbf{Efficiency and Model Size}} \\
    \cmidrule{2-6}\cmidrule{8-9}
     & mIoU(\%)$\uparrow$ & Recall(\%)$\uparrow$ & Precision(\%)$\uparrow$ & F1(\%)$\uparrow$ & sSR(\%)$\uparrow$ & & Inference Time (s) & Model Size (GB) \\
    \midrule
    RSMPNet~\cite{sun2024rsmpnet} & 26.49 & 30.09 & 53.99 & 37.15 & 0.00 && 0.016 & 0.44 \\
    SSCNav~\cite{liang2021sscnav} & \second{33.56} & 45.50 & 40.12 & \first{42.64} & 6.14 && 0.046 & \second{0.20} \\
    AOT-GAN~\cite{zeng2022aggregated} & 21.58 & 36.24 & \second{40.88} & 30.72 & 0.01 && \first{0.007} & \first{0.06} \\
    SGM~\cite{zhang2024imagine} & 25.88 & \first{57.06} & 29.56 & 35.43 & 21.88 && 0.013 & 1.52 \\
    FastMap-VQVAE & 26.55 & 35.60 & 31.09 & 32.61 & 5.35 && 0.012 & 0.42 \\
    FastMap w.o. ObMask & 29.32 & 40.00 & 36.01 & 37.90 & 4.56 && 0.011 & 0.41 \\
    FastMap (Ours) & \first{34.10} & \second{45.65} & \first{41.14} & \second{42.30} & \first{45.84} && \second{0.011} & 0.41 \\
    \bottomrule
  \end{tabular}
  }
\end{table*}

\subsection{Results}
\label{sec:exp_results}

\begin{figure}[t]
  \centering
  \begin{minipage}[b]{0.46\linewidth}
  \centering
  \resizebox{\linewidth}{!}{%
  \begin{tabular}{@{}l c c c@{}}
  \toprule
  Method & SR(\%)$\uparrow$ & SPL(\%)$\uparrow$ & DTS(m)$\downarrow$ \\
  \midrule
  Random          & 0.4  & 0.4  & 3.89 \\
  \midrule
  DD-PPO~\cite{wijmans2019dd}    & 15.0 & 10.7 & 3.24 \\
  EmbCLIP~\cite{khandelwal2022simple}  & 68.1 & 39.5 & 1.15 \\
  FBE~\cite{yamauchi1997frontier}       & 48.5 & 28.9 & 2.56 \\
  ANS~\cite{chaplot2020neural}        & 67.1 & 34.9 & 1.66 \\
  SemExp~\cite{chaplot2020object}     & 71.1 & 39.6 & 1.39 \\
  PONI~\cite{ramakrishnan2022poni}      & 73.6 & 41.0 & 1.25 \\
  3D-aware~\cite{zhang20233d}  & 74.5 & 42.1 & 1.16 \\
  SGM~\cite{zhang2024imagine}            & 75.4 & 39.3 & 1.26 \\
  NaviFormer~\cite{xie2025naviformer}            & 82.6 & 40.9 & \first{0.76} \\
  GOAL~\cite{li2025distilling}            & \second{83.5} & \first{44.2} & \second{0.83} \\
  \midrule
  FastMap (Ours) & \first{83.8} & \second{43.7} & 0.94 \\
  \bottomrule
  \end{tabular}}
  \captionof{table}{Object goal navigation comparison on Gibson. SR(\%) = Success Rate, SPL(\%) = Success weighted by Path Length, DTS(m) = Distance to Success.}
  \label{tab:comparison_navigation}
  \end{minipage}\hfill
  \begin{minipage}[b]{0.5\linewidth}
  \centering
  \includegraphics[width=\linewidth]{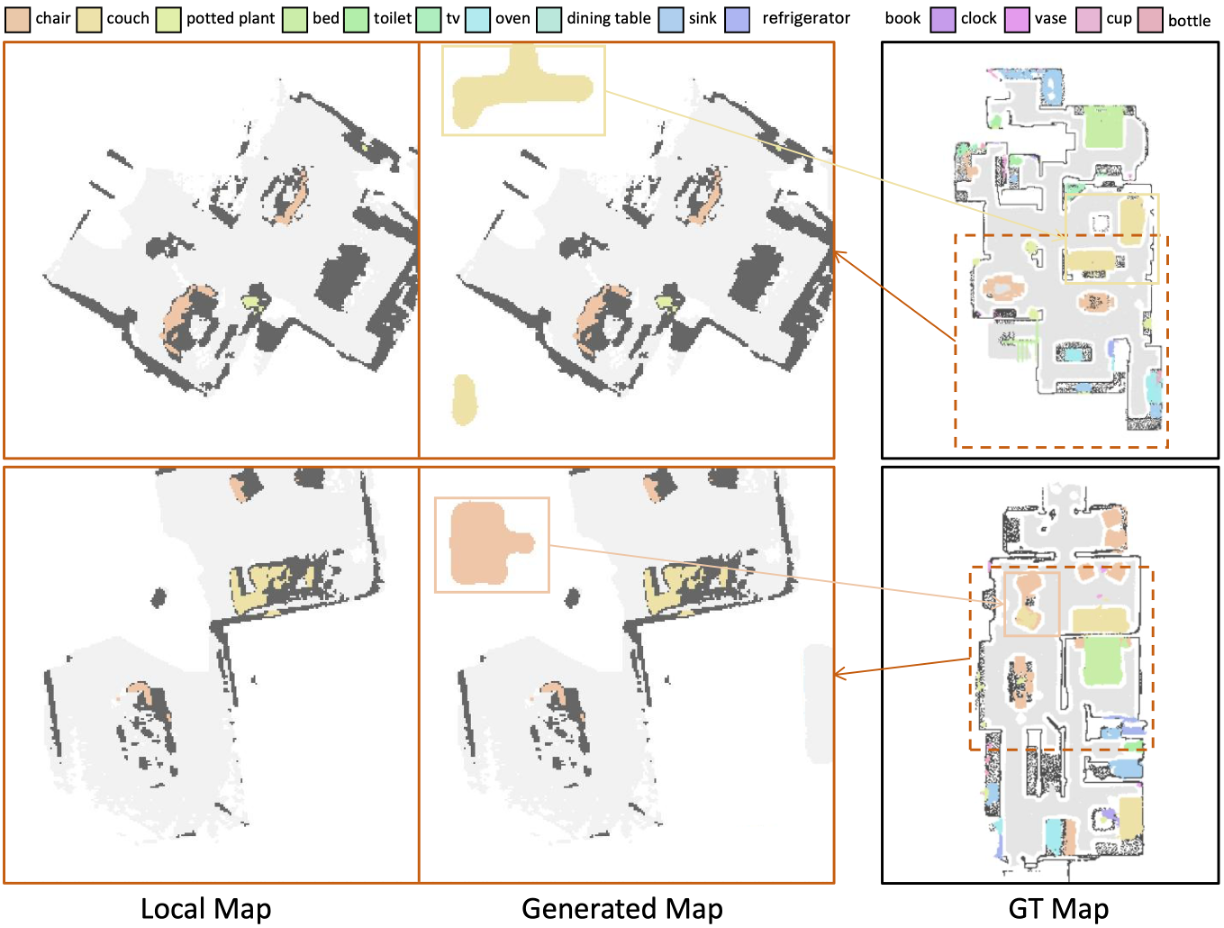}
  \caption{Object goal navigation results on Gibson~\cite{xia2018gibson}. From left to right: local map from the agent's observations, semantic map generated by our method, and ground-truth global map.}
  \label{fig:navigation}
  \end{minipage}
\end{figure}

\textbf{Map generation quality.} Under an object-aware masking protocol (we mask 50\% of the map: one whole object category plus random patches), FastMap attains the highest mIoU (34.10\%) in Table~\ref{tab:combined} and strong precision (41.14\%) and F1 (42.30\%) comparable to SSCNav. Its sSR reaches 45.84\%, more than doubling SGM's 21.88\%, reflecting recovery of fully masked target categories in unobserved regions.

\textbf{Qualitative comparison.} In Figure~\ref{fig:map_comparison}, SGM~\cite{zhang2024imagine} occasionally locates targets but its maps lack coherent structure, and SSCNav~\cite{liang2021sscnav} improves on it. FastMap-VQVAE and FastMap without object-aware masking produce plausible completions but miss precise localization, \rev{whereas the full FastMap reliably predicts object positions}.


\textbf{Efficiency and size.} Table~\ref{tab:combined} also compares inference time and size. \rev{All times are measured on a single NVIDIA RTX 4090 GPU with batch size one, averaged over the evaluation split, under the same protocol for all methods.} FastMap runs at 0.011 s per map with a compact 0.41 GB model, versus SGM's 1.52 GB. \rev{AOT-GAN~\cite{zeng2022aggregated} is faster/smaller and SSCNav~\cite{liang2021sscnav} smaller but much slower, yet both underperform in map quality; FastMap thus balances quality with a low footprint suited for deployment.}

\textbf{Object goal navigation.} In Table~\ref{tab:comparison_navigation}, semantic-exploration methods (SemExp, PONI, 3D-aware) reach $71$--$75\%$ SR, while recent transformer models push further: NaviFormer reaches 82.6\% SR with the lowest DTS (0.76 m) and GOAL 83.5\% SR with the best SPL (44.2\%). FastMap achieves the highest SR (83.8\%) with competitive 43.7\% SPL and 0.94 m DTS. \rev{No single method dominates all three metrics. We do not claim a uniformly superior system; rather, as a lightweight real-time module, FastMap matches far heavier systems and attains the best Success Rate.}

\begin{table}[t]
  \centering
  \caption{Ablation on quantization structure design choices on Gibson~\cite{xia2018gibson}.}
  \label{tab:ablation_vqvae}
  \resizebox{0.62\linewidth}{!}{%
    \begin{tabular}{lccc}
      \toprule
      Quant. & Code Dim & Codes/Bits & FID$\downarrow$ \\
      \midrule
      \multirow{3}{*}{VQVAE} & \multirow{3}{*}{16} & 256 & 1.74 \\
      & & 512 & 1.75 \\
      & & 1024 & \second{1.70} \\
      \midrule
      \multirow{3}{*}{BitVAE} & \multirow{3}{*}{--} & 8 & 1.75 \\
      & & 9 & 1.72 \\
      & & 10 & \first{1.65} \\
      \bottomrule
    \end{tabular}%
  }
\end{table}

\subsection{Ablation Study}
\label{sec:exp_ablation}

\renewcommand{\arraystretch}{1.0}
\begin{figure}[t]
  \centering
  \begin{minipage}[b]{0.57\linewidth}
    \centering
    \resizebox{\linewidth}{!}{
    \begin{tabular}{l|cc|ccccc}
    \hline
    Quant. & Codes/Bits & Mask & mIoU(\%) & Recall(\%) & Prec.(\%) & F1(\%) & sSR(\%) \\
    \hline
    \multirow{4}{*}{VQVAE}
      & 256 & R & 29.01 & 39.50 & 35.59 & 36.27 & 8.22 \\
     & 512 & R & 26.55 & 35.60 & 31.09 & 32.61 & 5.35 \\
     & 1024 & R & 26.73 & 36.75 & 32.90 & 33.44 & 9.50 \\
     & 1024 & O & 29.47 & 40.42 & 36.20 & 36.81 & \second{41.58} \\
    \hline
    \multirow{4}{*}{BitVAE}
      & 8  & R & 29.32 & 40.00 & 36.01 & 37.90 & 4.56 \\
     & 9  & R & \second{30.43} & \second{40.72} & \second{37.22} & \second{37.83} & 6.04 \\
     & 9  & O & \first{34.10} & \first{45.65} & \first{41.14} & \first{42.30} & \first{45.84} \\
     & 10 & R & 30.10 & 39.45 & 36.25 & 37.78 & 4.92 \\
    \hline
    \end{tabular}
    }
    \captionof{table}{Ablations of map generation quality on Gibson (ViT-B mask transformer). R = random masking, O = object-aware masking. ViT-B vs.\ ViT-L is reported in the text.}
    \label{tab:map_quality_overall}
  \end{minipage}\hfill
  \begin{minipage}[b]{0.4\linewidth}
    \centering
    \includegraphics[width=\linewidth]{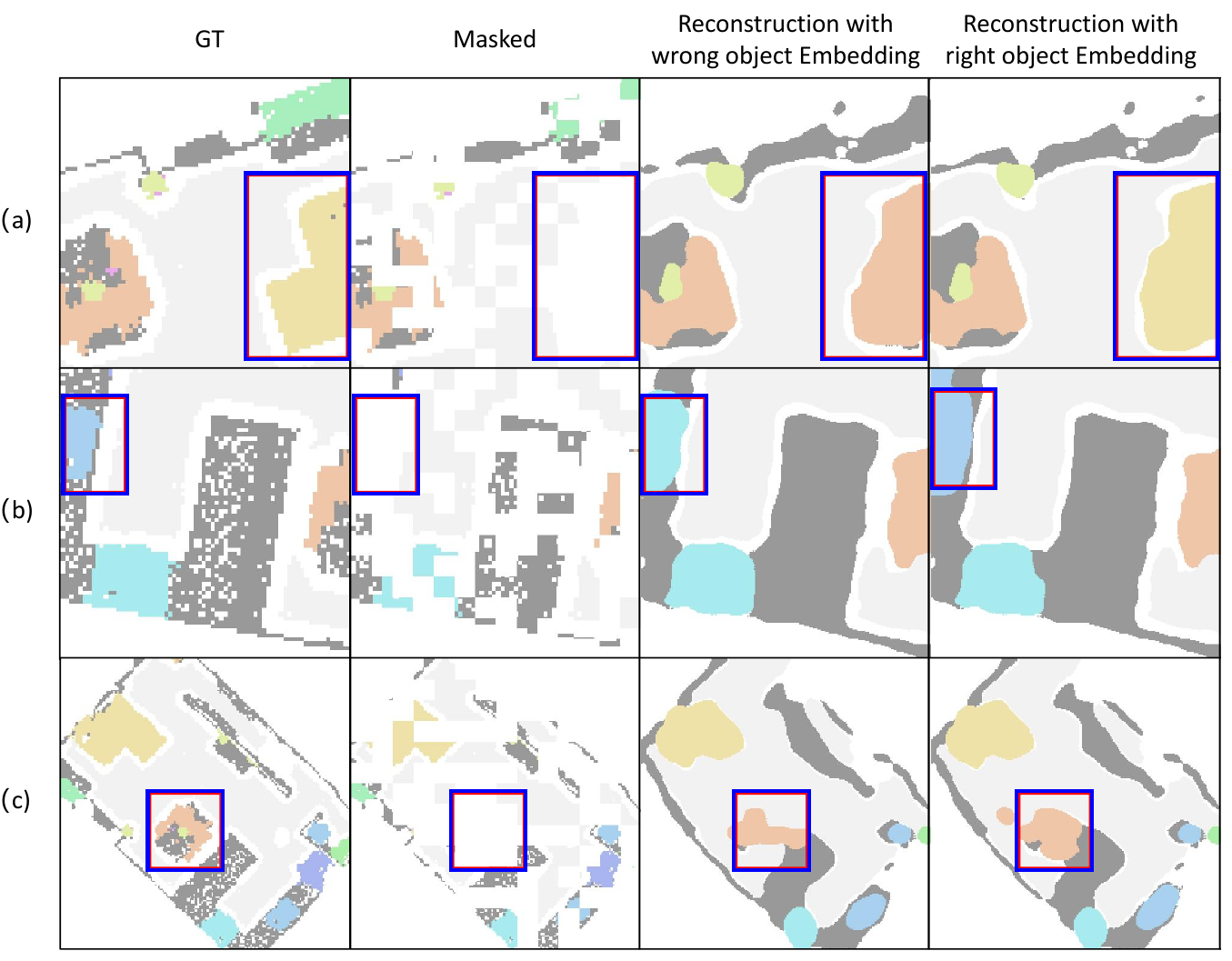}
    \captionof{figure}{Impact of correct/wrong object embeddings on map generation.}
    \label{fig:masking_strategy}
  \end{minipage}
\end{figure}

\textbf{VQVAE vs. BitVAE.} We compare VQVAE (codebook 256/512/1024, code dim 16) against BitVAE (8/9/10 bits). \rev{Here FID is the Fr\'echet Inception Distance~\cite{heusel2017gans} (lower is better); since our maps are categorical, we compute it over tokenizer reconstructions with a fixed backbone and treat it only as a \emph{relative} quality indicator, consistent with the mIoU/F1 of Table~\ref{tab:map_quality_overall}.} In Table~\ref{tab:ablation_vqvae}, BitVAE with 10 bits attains the lowest FID, confirming the value of direct bit-wise encoding. \rev{Downstream (VQVAE with 512 codes vs.\ BitVAE with 9 bits),} BitVAE also gives 10\% higher masked-token accuracy, showing its binary representations better capture inter-object relations.

\textbf{Mask transformer model size.} \rev{Enlarging the mask transformer from ViT-B to ViT-L barely changes map-generation quality (e.g., BitVAE 9-bit mIoU is 30.43\% for both), suggesting a smaller architecture already suffices, so we adopt ViT-B.}

\textbf{Masking strategy.} Object-aware masking clearly beats random masking: for both VQVAE- and BitVAE-based transformers it gives 3-4\% better map metrics and up to 10$\times$ higher sSR, \rev{since masking target objects while supplying their embeddings teaches the model how embeddings map to tokens}. Figure~\ref{fig:masking_strategy} shows correct embeddings improve localization (rows a-b), while the model stays robust to wrong ones given enough context (row c).

\section{Conclusion}
We presented FastMap, an efficient framework for top-down semantic map completion. Combining a lookup-free BitVAE tokenizer, a lightweight masked transformer, and target-conditioned object-aware masking, it attains strong quality on Gibson (34.10\% mIoU, 45.84\% sSR) at real-time speed (0.011\,s/map, 0.41\,GB) and raises ObjectNav success to 83.8\% within a navigation pipeline. \rev{\textbf{Limitations and future work.} Our study has two main limitations. First, all experiments use Gibson; single-dataset evaluation limits generalizability, so validating FastMap on more datasets (e.g., MP3D~\cite{chang2017matterport3d} and HM3D) and on a real robot is an important next step. Second, our maps come from simulation, so we do not yet characterize robustness to real-deployment perturbations such as perception noise, odometry drift, and map misalignment.}

\section{Acknowledgements}
Authors appreciate the support provided by the NYUAD Center for Artificial Intelligence and Robotics (CAIR), funded by Tamkeen under the NYUAD Research Institute Award CG010.

{
    \footnotesize
    \bibliographystyle{IEEEtran}
    \bibliography{ref}
}

\end{document}